\ifdefined\pdfoutput\pdfoutput=1\fi 

\documentclass[10pt, a4paper]{article}
\usepackage[]{lrec-coling2024} 
\usepackage{multirow}
\usepackage{natbib}
\usepackage{multibib}
\usepackage{siunitx}
\usepackage{bm}
\usepackage{amsmath}
\usepackage{amsthm}
\usepackage{enumitem}
\usepackage{amssymb}
\usepackage{amsfonts}
\usepackage{booktabs}
\usepackage[ruled,vlined]{algorithm2e}
\usepackage{physics}
\usepackage{bbm}

\makeatletter
\def\@mb@citenamelist{cite,citep,citet,citealp,citealt,citepalias,citetalias}
\makeatother
\newcites{languageresource}{~}

\usepackage{graphicx}
\usepackage{tabularx}
\usepackage{soul}
\usepackage{titlesec}
\titleformat{\section}{\normalfont\large\bfseries\center}{\thesection.}{1em}{}
\titleformat{\subsection}{\normalfont\SmallTitleFont\bfseries\raggedright}{\thesubsection.}{1em}{}
\titleformat{\subsubsection}{\normalfont\normalsize\bfseries\raggedright}{\thesubsubsection.}{1em}{}
\renewcommand\thesection{\arabic{section}}
\renewcommand\thesubsection{\thesection.\arabic{subsection}}
\renewcommand\thesubsubsection{\thesubsection.\arabic{subsubsection}}

\usepackage{xcolor}

\usepackage{hyperref}
 \definecolor{darkblue}{rgb}{0, 0, 0.5}
  \hypersetup{colorlinks=true, citecolor=darkblue, linkcolor=darkblue, urlcolor=darkblue}

\usepackage{xstring}
\usepackage{natbib}
\usepackage{amssymb}
\usepackage{amsmath}

\usepackage{bm}
\usepackage{bbm}
\usepackage{graphicx}
\usepackage{color}

\def\by{\mathrm{y}}

\title{LLMR: Knowledge Distillation with a Large Language Model-Induced Reward}

\name{Dongheng Li$^{1}$, Yongchang Hao$^{1}$, Lili Mou$^{1,2}$} 

\address{$^1$Dept.~Computing Science \& Alberta Machine Intelligence Institute (Amii), 
University of Alberta\\$^2$Canada CIFAR AI Chair, Amii\\\texttt{\{dongheng,yongcha1\}@ualberta.ca,
  doublepower.mou@gmail.com} \\}

\abstract{
Large language models have become increasingly popular and demonstrated remarkable performance in various natural language processing (NLP) tasks. However, these models are typically computationally expensive and difficult to be deployed in resource-constrained environments. In this paper, we propose LLMR,  a novel knowledge distillation (KD) method based on a reward function induced from large language models. We conducted experiments on multiple datasets in the dialogue generation and summarization tasks. Empirical results demonstrate that our LLMR approach consistently outperforms traditional KD methods on different tasks and datasets.}

\begin{document}

\maketitleabstract

\section{Introduction}\label{sec:intro}
Large language models (LLMs) have achieved remarkable performance in various text generation tasks, such as summarization~\citep{ahmed2022fewshot,nair2023dera} and dialogue systems~\citep{deng2023prompting,cao-etal-2020-pretrained}. Moreover, this can be accomplished in a zero-shot manner, that is, a user enters a natural language prompt (e.g., ``\textit{Summarize the following text}'') and the LLM will generate a desired output for the task~\cite{brown2020language}.

However, LLMs also present significant challenges. For example, the GPT-3 model has 175 billion parameters, which is resource-intensive, requiring significant computing power and memory. This might hinder real-world applications in resource-constrained environments.

Therefore, knowledge distillation~\cite[KD;][]{hinton2015distilling} becomes an increasingly important research direction for LLMs~\cite{gu2023knowledge,wu2023laminilm,hsieh2023distilling}, where the goal is to transfer the knowledge of LLM (called a \textit{teacher}) to a smaller and more efficient model (called a \textit{student}). Conventionally, this is accomplished by training the student from the teacher's predicted sentences or distributions~\cite{kim-rush-2016-sequence}. However, it has inherent limitations:
during training, the student learns to predict the next word based on the teacher's previous predictions, whereas during inference, the student has to do so based on its own previous predictions. Such a discrepancy is known as \textit{exposure bias}, and often leads to a performance degradation~\citep{chiang-chen-2021-relating,ranzato2016sequence}.

In this paper, we propose a novel knowledge distilling method, based on reinforcement learning with a Large Language Model-induced Reward (dubbed LLMR). 
Instead of directly training from LLM's output, we first induce a $q$-value function from the LLM's policy (predicted probabilities) based on a widely adopted assumption~\citep{chan2021scalable,ramachandran2007bayesian,ziebart2008maximum}, and then further induce a reward function based on the Bellman optimality equation~\cite{policy_gradient}; this process follows our recent theoretical analysis between policies and rewards~\cite{NEURIPS2022_51ae7d9d}. The induced reward function is subsequently used to distill LLM's knowledge into the student, achieved by sampling a candidate sequence from the student-predicted distributions and evaluating it with the LLM-induced reward for policy gradient learning~\cite{REINFORCE}. In this way, our proposed LLMR distilling approach allows the student model to explore on its own during KD in a reinforcement learning (RL)~fashion, thus alleviating the exposure bias problem.

We conducted experiments on two text generation tasks: dialogue generation and text summarization. Empirical results show that our LLMR approach largely outperforms traditional KD based on cross-entropy loss. We further quantitatively analyzed the exposure bias of the student models, verifying that RL indeed alleviates exposure bias arising during the KD process.\footnote{Our code is released as a GitHub repo: \url{https://github.com/MANGA-UOFA/Prompt-LLMR}}

\section{Related Work}
Knowledge distillation (KD) is effective in reducing the computing and memory demands of large neural networks while retaining high performance. Common KD approaches include matching output distributions~\cite{hinton2015distilling,wu2023laminilm} and matching intermediate-layer representations~\cite{Romero15-iclr,distillandquant,sun2019pkd}. 

KD has been applied to the sequence level for distilling text generation models~\cite{kim-rush-2016-sequence, ebbs} and autoregressive language models~\cite{west-etal-2022-symbolic}. Typically, the student learns from the teacher step by step with a cross-entropy loss, but such an approach may suffer from exposure bias~\cite{ranzato2016sequence}. Researchers have proposed reverse Kullback--Leibler~\cite{tu-etal-2020-engine,gu2023knowledge} and generalized $f$-divergence~\cite{wen-etal-2023-f} losses, which involve student sampling but still follow the spirit of traditional KD pushing the student's distribution to the teacher's step by step. In our LLMR method, on the other hand, the teacher only scores a student-sampled sequence, which allows more exploration during the KD process.

Reinforcement learning (RL) has been widely used for text generation, especially for alleviating exposure bias~\citep{ranzato2016sequence,gu2023knowledge}. A key design choice is the reward function, which in previous work is often given by task heuristics with groundtruth sequences~\cite{sokolov2016stochastic,gold} or trained reward models~\cite{bahdanau2016actor,rlSum}. Our LLMR method follows previous theoretical work~\cite{NEURIPS2022_51ae7d9d}, but directly induces a reward function from a pretrained LLM in a principled and task-agnostic manner.

\section{Approach}

\textbf{Problem Formulation.}
Knowledge distillation (KD) aims to transfer the knowledge of a teacher model to a student. Although the student can solely learn a task from a parallel corpus $D_p = \{(\mathbf x^{(i)},\mathbf y^{(i)})\}^M_{i=1}$, it is argued that the teacher's predicted distribution contains more knowledge than an annotated label $\mathbf y$~\cite{hinton2015distilling}.
\citet{kim-rush-2016-sequence} propose SeqKD and minimize a Kullbuck--Leibler loss, equivalent to minimizing a cross-entropy loss, at the sequence level between a teacher $p$ and a student $q_\theta$ by $J_{\text{SeqKD}} = \mathbb E_{\mathbf y \sim p} \left[\log\frac{p(\mathbf y|\mathbf x)}{q_\theta(\mathbf y |\mathbf x)}\right]
$.
In practice, the expectation over the sentence space is intractable. To tackle this, they use a hard sequence ${\mathbf y}$ generated by beam search on the teacher model as an approximation: $\hat J_{\text{SeqKD}} = - \log q_{\theta}({\mathbf y}|\mathbf x)$.

\begin{figure}[!t]\centering
        \includegraphics[width=1.0\linewidth]{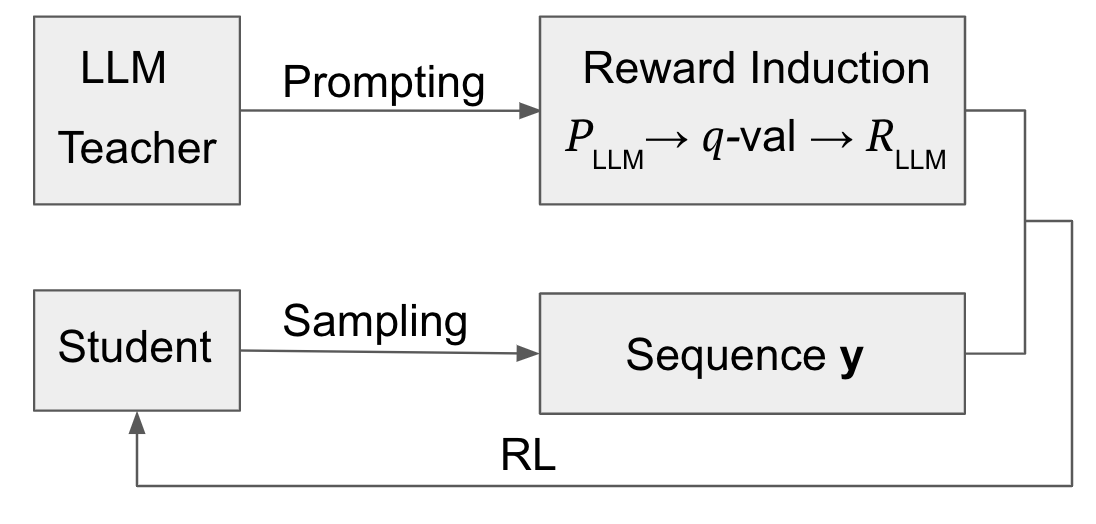}
        \caption{Overview of the approach.}
\label{fig:overview}
\end{figure}

In our work shown in Figure~\ref{fig:overview}, we prompt a large language model (LLM) and treat it as the teacher. However, we do not follow the common KD that minimizes the divergence between LLM's probability $p_\text{LLM}$ and the student $q_\theta$. Instead, we propose to induce a reward function $R_\text{LLM}$ from $p_\text{LLM}$ and adopt reinforcement learning for KD with objective:
\begin{align}\label{eqn:objective}
\operatorname{maximize}_{\theta} \mathbb E_{\mathbf y\sim q_\theta}[R_\text{LLM}(\mathbf y
)]
\end{align}
Our approach alleviates the exposure bias problem~\citep{chiang-chen-2021-relating,ranzato2016sequence} in traditional KD, where the student is fed with the teacher's predicted prefix during training, but only has access to its own partial prediction during inference. By contrast, our RL-based KD allows the student  to explore with its own predicted sequence, shown by $\mathbf y\sim q_\theta$ in (\ref{eqn:objective}), which bridges the gap between training and inference.

In the rest of this section, we will introduce the reward $R_\text{LLM}$ and the optimization of  (\ref{eqn:objective}) in detail.

\textbf{Inducing Reward from LLMs.}
We propose to induce a reward function from large language models (LLMs) for RL-based KD, inspired by the theoretical analysis that links policies (predicted probabilities) and reward functions~\cite{NEURIPS2022_51ae7d9d}. In our work, we design an intuitive prompt to obtain the LLM's policy for reward induction.

Consider a task $\mathcal T$ and an input sentence $\mathbf x$. We formulate a prompt as $\texttt{pmt}_{\mathcal T}(\mathbf x)$. In fact, the prompt depends on the task of interest, and in our experiments, two common text generation tasks are considered: summarization~\citep{ahmed2022fewshot,nair2023dera} and dialogue generation~\citep{deng2023prompting,cao-etal-2020-pretrained}. Our prompts are
\begin{align}\nonumber
    \texttt{pmt}_\text{sum}(\mathbf x)&\equiv\text{``Summarize [ $\mathbf x$ ]:"}\\\nonumber
    \texttt{pmt}_\text{dialog}(\mathbf x)&\equiv\text{``The dialogue response of [ $\mathbf x$ ] is:"}
\end{align}
where $\mathbf{x}$ is the original input sentence and the square brackets are delimiters specifying the input boundaries. 

Given a candidate output $\mathbf y=(\mathrm y_1,\cdots, \mathrm y_T)$, our goal is to induce a reward function $R_\text{LLM}(\mathbf y)$ that evaluates the ``goodness'' of $\mathbf y$. This requires modeling text generation as a Markov decision process (MDP), where an action is the prediction of the next word and a state is the partially predicted sequence in addition to the prompt. The state transition is a deterministic process that simply appends the newly generated word to the previous state.

Our reward induction starts by querying an LLM in a step-by-step fashion to obtain the next word probability $p_\text{LLM}(\mathrm y_t|\mathbf y_{<t}, \texttt{pmt}_{\mathcal T}(\mathbf x))$. Notice that we do not let the LLM generate outputs during RL-based KD, but the prefix $\mathbf y_{<t}$ and the next word $\mathrm y_t$ are from the student-sampled sequence. The role of LLM is to predict its probability and to induce a reward for $\mathbf y$.

\newcommand{\qvalue}{q\texttt{-}\mathtt{val}}
With the next-word probability, we are able to induce a $q$-value function for step $t$, which indicates the goodness of an action, i.e., the word $\mathrm y_t$, at the state $(\mathbf y_{<t}, \texttt{pmt}_{\mathcal T}(\mathbf x))$. The $q$-value induction process is based on the common assumption~\citep{chan2021scalable,ramachandran2007bayesian,ziebart2008maximum} that an action is taken stochastically based on a Boltzmann distribution induced by $q$-values:
\begin{align}
\hspace{-.3cm}\resizebox{0.97\linewidth}{!}{
    $\begin{aligned}\label{eq:qvalue}
    &p_\text{LLM}(\mathrm y_t|\mathbf y_{<t}, \texttt{pmt}_{\mathcal T}(\mathbf x)) =\frac{\exp\{\qvalue(\mathrm y_t; \mathbf y_{<t})\}}{\sum_{\mathrm y'} \exp\{\qvalue(\mathrm y'; \mathbf y_{<t})\}}\end{aligned}$
}\hspace{-.3cm}
\end{align}
where the $q$-value function also depends on $\texttt{pmt}_{\mathcal T}(\mathbf x)$ but is omitted for simplicity.

In other words, the assumption implies that a higher-valued action will be taken with a larger probability, which makes much sense in practice. Moreover, the resemblance between \eqref{eq:qvalue} and a softmax function suggests that we may directly take the LLM's logit (pre-softmax value) $f_\text{LLM}$ as the $q$-value in the MDP modeling.

The final step of reward induction is based on Bellman optimality~\citep{degris2012off,sutton2018reinforcement}, which derives an optimal $q$-value function from a reward. We follow the practice of inverse reinforcement learning~\citep{ramachandran2007bayesian,ziebart2008maximum,chan2021scalable} and use Bellman optimality in an opposite way to derive a reward $R_\text{LLM}$ from the $q$-value function in~\eqref{eq:qvalue}:
\begin{align}\label{eq:reward}
\resizebox{0.8\linewidth}{!}{
    $\begin{aligned}
     R_\text{LLM}(\mathrm{y}_t; \mathbf{y}_{<t}) =&\qvalue(\mathrm{y}_t; \mathbf{y}_{<t}) \\
    &-\max\nolimits_{\mathrm y'} \qvalue(\mathrm{y}';\mathbf{y}_{<t+1})
    \end{aligned}$
}
\end{align}

In this way, our derived reward $R_\text{LLM}$ evaluates the appropriateness of a word $\mathrm y_t$ at every step given its context $\mathbf y_{<t}$. That is to say, such a reward function is dense as opposed to various other heuristic rewards (e.g., BLEU scores) that only come at the end of a sequence~\cite{wu2017sequence}. The overall reward induction process follows
our previous work~\cite{NEURIPS2022_51ae7d9d}, but this paper extends it to a new scenario.  \citet{NEURIPS2022_51ae7d9d} train a sequence-to-sequence network in a supervised manner on a parallel corpus and perform semi-supervised learning on non-parallel corpora. Our paper shows that a reward function can be derived directly from a pretrained LLM and helps various text generation tasks, which is a new insight, especially in the LLM era.

\begin{table*}[t]
\centering
\resizebox{0.9\textwidth}{!}{%
\begin{tabular}{|cll|cc|cc|ccc|}
\hline
\multicolumn{3}{|c|}{} &
  \multicolumn{2}{c|}{DailyDialog} &
  \multicolumn{2}{c|}{OpenSubtitles} &
  \multicolumn{3}{c|}{CNN/DailyMail} \\ \cline{4-10} 
\multicolumn{3}{|c|}{\raisebox{1.1ex}{Model}} &
  \multicolumn{1}{c}{BLEU2} &
  \multicolumn{1}{c|}{BLEU4} &
  \multicolumn{1}{c}{BLEU2} &
  \multicolumn{1}{c|}{BLEU4} &
  \multicolumn{1}{c}{ROUGE-1} &
  \multicolumn{1}{c}{ROUGE-2} &
  \multicolumn{1}{c|}{ROUGE-L} \\ \hline
1 & \multicolumn{2}{|l|}{Prompting Teacher} &
  5.57 &
  1.49 &
  4.67 &
  1.51 &
  36.16 &
  14.99 &
  24.05 \\ \hline
2 & \multicolumn{2}{|l|}{Prompting Student} &
  1.35 &
  0.31 &
  1.21 &
  0.25 &
  21.23 &
  6.73 &
  17.88 \\ \hline
3 & \multicolumn{1}{|l|}{\multirow{5}{*}{\shortstack{Distilled\\ Students}}} &
  SeqKD &
  6.19 &
  1.71 &
  3.87&
  1.35 &
  35.46 &
  14.52 &
  23.68 \\
4 & \multicolumn{1}{|l|}{} &
  KL &
  5.03 &
  1.40 &
  3.84 &
  1.33 &
  34.11 &
  14.21 &
  22.83 \\
5 & \multicolumn{1}{|l|}{} &
  RKL &
  5.02 &
  1.29 &
  4.12 &
  1.36 &
  32.07 &
  13.77 &
  22.87 \\
6 & \multicolumn{1}{|l|}{} &
  JS &
  6.60 &
  1.73 &
  3.64 &
  0.87 &
  35.88 &
  14.72 &
  23.97 \\
7 & \multicolumn{1}{|c|}{} &
  Our LLMR &
  \textbf{7.00} &
  \textbf{1.88} &
  \textbf{5.13} &
  \textbf{1.85} &
  \textbf{36.42} &
  \textbf{15.21} &
  \textbf{24.83} \\
 \hline
\end{tabular}%
}\vspace{-.2cm}
\caption{Main results on dialogue generation and summarization tasks.}\vspace{-.2cm}
\label{tab:combined}
\end{table*}
\textbf{Reinforcement Learning-Based KD.}
Our derived reward function allows us to perform reinforcement learning (RL) for KD. Specifically, a sequence is sampled from the student's prediction, given by $\mathbf y\sim q_\theta$. Then, each word $\mathrm y_t$ in $\mathbf y$ is evaluated by the induced reward function \eqref{eq:reward}, and our total reward of the sequence is\begin{align}
R_\text{LLM}(\mathbf y)=\sum\nolimits_t R_\text{LLM}(\mathrm y_t;\mathbf y_{<t})
\end{align}
which is our objective to maximize, as shown in Eqn.~\eqref{eqn:objective}.

Since the parameter $\theta$ occurs during the sampling process, the gradient cannot be obtained by backpropagation, and RL is required to train $\theta$ in a trial-and-error manner. In NLP, the REINFORCE method is commonly used~\citep{ranzato2016sequence,abs-1909-01150}, where the gradient is given by
\begin{align}\nonumber
\resizebox{1.1\linewidth}{!}{
    $\begin{aligned}
        \nabla_{\theta}\operatorname*{\mathbb{E}}\limits_{\pi_\theta}\!\left[\sum_{t} R_\text{LLM}(\mathrm y_t;\mathbf y_{<t})\right] &=
        \operatorname*{\mathbb{E}}\limits_{\pi_\theta}\!\left[\sum_{t} G_t(\mathbf y) \nabla_{\theta}\log \pi_\theta(\mathrm y_t;\mathbf y_{<t})\right]
    \end{aligned}$
}
\end{align}where $G_t(\mathbf y)$ is known as the gain in the RL literature, being the accumulated reward from step $t$, given by
$G_t(\mathbf y) := \sum_{\tau\ge t} R_\text{LLM}(\mathrm y_\tau;\mathbf y_{<\tau})$.

Overall, our RL-based KD differs from traditional sequence-level KD, where the teacher teaches unilaterally with its own prediction, i.e., \( \mathbf{y} \sim p_\text{LLM} \). Instead, we allow the student to generate its own prediction, and the LLM teaches by evaluating the ``goodness'' of the student's output. In this way, our approach alleviates the exposure bias problem, as the student is aware of its own partial prediction during training. Compared with classic RL-based text generation, we do not require heuristically designed reward functions~\citep{bahdanau2016actor,shen2015minimum} or human feedback reward models~\citep{tlhf,finehf}.

\section{Experiments}
\textbf{Setups.} 
We evaluated our approach on two text generation tasks with three datasets: DailyDialog~\cite{li2017dailydialog} and OpenSubtitles~\cite{tiedemann2009news} for dialogue generation, as well as CNN/DailyMail~\citep{see-etal-2017-get,HermannKGEKSB15} for summarization. In particular, dialogue datasets tend to have sample-overlapping issues between training and test sets, and we used the cleaned version~\cite{wen2022empirical} for rigorous experimentation.

Our teacher was a T0-3B model~\cite{sanh2022multitask} and the student was T5-Base with 220 million parameters~\cite{raffel2019exploring}. Since our RL-based KD requires meaningful sampling from the student, we performed pre-distillation by the standard cross-entropy loss~\cite{kim-rush-2016-sequence}, which is common in KD research~\citep{wen-etal-2023-f,shleifer2020pre} and shows our method provides add-on improvement. 

It should be emphasized that our work addresses unsupervised KD, where the training process only used unlabeled input sentences without groundtruth references. During validation and test phases, the ground truths were used in the standard evaluation metrics: BLEU~\cite{papineni2002bleu} for dialogue generation and ROUGE~\cite{lin2004rouge} for summarization.

\textbf{Main Results.} Table~\ref{tab:combined} 
presents the performance of our model and baselines. As seen,  the teacher model (Row~1) achieves decent performance in these tasks. The results are slightly lower than, or comparable to, those of supervised methods reported in previous literature, for example, 8.96 BLEU2 for DailyDialog~\cite{NEURIPS2022_51ae7d9d} and 39.5 ROUGE-1 for CNN/DailyMail~\cite{vaswani2017attention}. This is understandable because our teacher is directly prompted for the tasks without finetuning. On the other hand, prompting the student (Row~2) does not yield meaningful performance, which is consistent with the findings of the scaling effect~\citep{kim-rush-2016-sequence,hinton2015distilling,wen-etal-2023-f}. The strong teacher and weak student jointly set up a reasonable foundation for our distillation research.

Rows 3--7 present the performance of different distilling methods, showing that KD can indeed transfer the teacher's knowledge into the student. Among different KD methods, SeqKD~\cite{kim-rush-2016-sequence} employs hard samples to train the student, and achieves close performance to the teacher; in particular, it surpasses the teacher on DailyDialog, which can be interpreted by smoothing the noise of the teacher (an un-finetuned prompting system). We also experimented with soft distillation based on various $f$-divergence functions, including Kullback--Leibler (KL), Reverse KL (RKL), and Jenson--Shannon (JS) divergences~\cite{wen-etal-2023-f}. As seen, the results are not fully consistent, although JS tends to perform better in general. 

Our LLMR (Row~7) performs reinforcement learning based on a reward function induced from the teacher model. It achieves superior performance across all the metrics and datasets, consistently demonstrating the effectiveness of our approach. 
\begin{table}[t]
\centering

\resizebox{\linewidth}{!}{
\begin{tabular}{lcccccc}
\toprule
& \multicolumn{2}{c}{DailyDialog} & \multicolumn{2}{c}{OpenSubtitles} & \multicolumn{2}{c}{CNN/DailyMail\!\!} \\
\cmidrule(lr){2-3} \cmidrule(lr){4-5} \cmidrule(lr){6-7}
\!\!Model & Dist1 & Dist2 & Dist1 & Dist2 & Dist1 & Dist2\!\! \\
\midrule
\!\!SeqKD & 4.93 & 27.37 & 4.78 & 23.15 & 3.86 & 33.59\!\! \\
\!\!KL & 4.76 & 26.77 & 4.99 & 24.00 & 3.76 & 33.59\!\! \\
\!\!RKL & 5.76 & 29.01 & 5.38 & 23.72 & 4.07 & 32.27\!\! \\
\!\!JS & 5.84 & 32.25 & 4.44 & 19.21 & 3.83 & 31.47\!\! \\
\!\!Our LLMR\!\!\!\!\! & \textbf{6.02} & \textbf{34.83} & \textbf{5.82} & \textbf{27.21} & \textbf{4.20} & \textbf{35.38}\!\! \\
\bottomrule
\end{tabular}}\vspace{-.2cm}
\caption{Distinct $n$-gram (Dist$n$) scores.}\vspace{-.2cm}\label{tab:diverse_table}
\end{table}

\textbf{Diversity Analysis.} The diversity of output text is considered an important aspect of text generation systems~\citep{li-etal-2016-diversity,eqem}. We evaluated the diversity of competing models by the standard distinct $n$-gram measures~\citep{li-etal-2016-diversity,gold,halu_survey}, given by 
\begin{align*}\label{eq:diveranly}
    \text{Distinct-$n$} = \frac{\text{Number of unique $n$-grams } }{\text{Total number of $n$-grams }}
\end{align*}

As seen in Table~\ref{tab:diverse_table}, the KL loss achieves low distinct scores, which is consistent with previous evidence that the KL training makes the model generate dull and short utterances~\cite{responsesmou,eqem}. By contrast, our LLMR yields much higher distinct scores, which verifies that our RL mechanism allows the model to explore different regions of the sentence space, leading to much more diverse output. 

\begin{figure}[!t]\centering
        \includegraphics[width=1.0\linewidth]{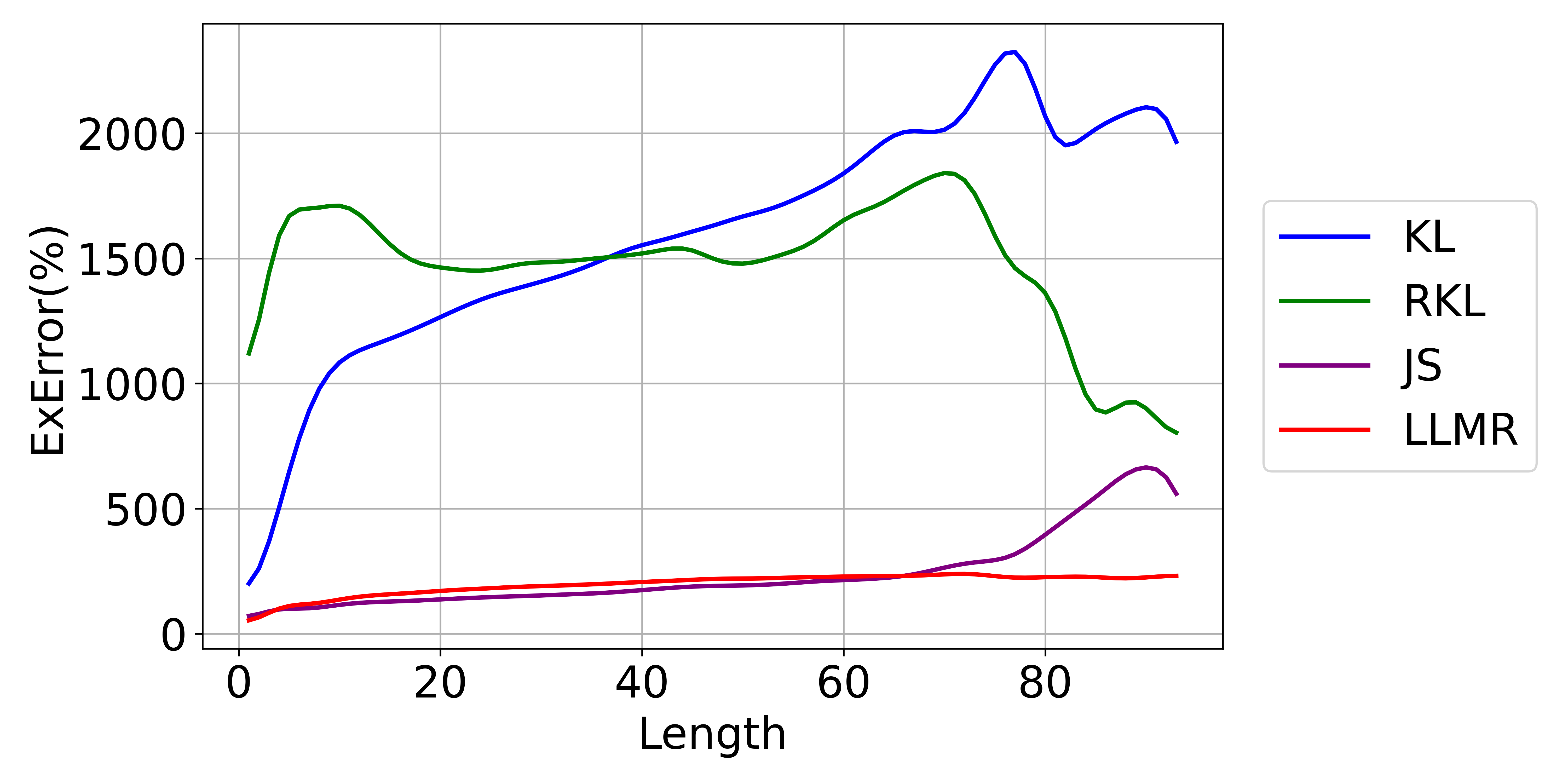}
        \caption{The averaged excess error (ExError) with respect to sequence length of different models on DailyDialog.}
\label{fig:exerror}
\end{figure}

\textbf{Exposure Bias Analysis.} As mentioned in \S\ref{sec:intro}, our LLMR adopts RL and is supposed to alleviate exposure bias during KD. We quantify the amount of exposure bias by adapting a recently established measure, Excess Error Percentage~\citep[ExError\%,][]{eb_measure}. In our scenario, ExError\% is defined by

\begin{equation*}
    \text{ExError\%} (l)= \frac{D_s(l) - D_t(l)}{D_t(l)} \times 100\%
\end{equation*}
Here, \(D_{\color{red}s}(l)\) stands for the accumulated Kullback--Leibler (KL) divergence between the teacher and student, when  the models follow the {\color{red}student}'s trajectory up to the $(t-1)$th step:
\begin{equation*}
\begin{split}
    D_{\color{red}s}(l) = \sum_{t=1}^T \mathbb{E}_{\substack{\by_{<t} \sim {\color{red}q_{\theta}}(\cdot | \mathrm{x}) \\ \mathrm y_t \sim p(\cdot | \mathbf{y}_{<t}, \mathrm{x})}} \Bigg[ \log \frac{p(\mathrm y_t | \mathbf{y}_{<t}, \mathrm{x})}{q_{\theta}( \mathrm y_t | \mathbf{y}_{<t}, \mathrm{x})} \Bigg]
\end{split}
\end{equation*}
whereas  \(D_{\color{blue}t}(l)\) is the KL divergence when the models follow the {\color{blue}teacher}'s trajectory up to the $(t-1)$th step:
\begin{equation*}
\begin{split}
    D_{\color{blue}t}(l) = \sum_{t=1}^T \mathbb{E}_{\substack{\by_{<t} \sim {\color{blue}p}(\cdot | \mathrm{x}) \\ \mathrm{y}_t \sim p(\cdot | \mathbf{y}_{<t}, \mathrm{x})}} \Bigg[ \log \frac{p(\mathrm{y}_t | \mathbf{y}_{<t}, \mathrm{x})}{q_{\theta}(\mathrm{y}_t | \mathbf{y}_{<t}, \mathrm{x})} \Bigg]
\end{split}
\end{equation*}
Overall, ExError\% measures the percentage of excess error when the models follow the {\color{red}student}'s trajectory, compared with following the {\color{blue}teacher}'s trajectory. Typically, ExError\% is positive and a higher value indicates more exposure bias. It can go over 100\% because the KL divergence is not upper bounded.

As seen in Figure~\ref{fig:exerror}, KL- and RKL-based KD methods yield high exposure bias, which is expected as the KL and RKL divergence functions are asymmetric and do not push the student to the teacher well. The JS divergence is symmetric and JS-based KD requires both teacher and student samplings. Its ExError\% remains low at the beginning, but grows when the sequence becomes longer. Our LLMR approach employs RL training and achieves low ExError\% throughout different lengths. The experiment confirms our approach alleviates exposure bias and explains the performance improvement in main results.

\section{Conclusion} 
In this paper, we propose a novel knowledge distillation method, called LLMR, based on a large language model-induced reward function. Experiments on dialogue generation and text summarization show that our approach outperforms previous KD methods in terms of various metrics. We also conducted a detailed analysis to verify that our reinforcement learning-based method indeed alleviates the exposure bias problem present in common KD approaches.

\section{Acknowledgments}
We thank all reviewers and chairs for their
valuable comments. The research is supported in
part by the Natural Sciences and Engineering Research Council of Canada (NSERC) under Grant
No. RGPIN2020-04465, an Alberta Innovates Project, the Amii Fellow Program,
the Canada CIFAR AI Chair Program, a UAHJIC
project, a donation from DeepMind, and the Digital
Research Alliance of Canada (alliancecan.ca).

\section{Bibliographical References}\label{sec:reference}

\vspace{-.6cm}
\bibliography{custom}
\bibliographystyle{lrec_natbib}

\end{document}